\documentclass{article} 
\usepackage{iclr2021_conference,times}

\usepackage{amsmath,amsfonts,bm}

\def\eqref#1{equation~\ref{#1}}

\def\1{\bm{1}}

\DeclareMathAlphabet{\mathsfit}{\encodingdefault}{\sfdefault}{m}{sl}
\SetMathAlphabet{\mathsfit}{bold}{\encodingdefault}{\sfdefault}{bx}{n}

\usepackage{hyperref}
\usepackage{url}
\usepackage[pdftex]{graphicx}
\usepackage{multirow}
\usepackage{subfigure}
\usepackage{makecell}
\usepackage{amsmath}
\usepackage{booktabs}
\usepackage{enumitem}
\usepackage{mathtools, nccmath}
\usepackage{setspace}
\usepackage{multirow}
\newcommand{\etal}{\textit{et al.}}

\makeatletter
\newcommand{\printfnsymbol}[1]{\textsuperscript{\@fnsymbol{#1}}
}
\makeatother

\newlist{myitemize}{enumerate}{10}
\setlist[myitemize]{label*=\arabic*.,nosep,leftmargin=*}

\title{\centerline{Learning Graph Normalization for} 
 \centerline{Graph Neural Networks}}

\author{Yihao Chen\thanks{Equal Contribution.}, Xin Tang\printfnsymbol{1}, Xianbiao Qi\printfnsymbol{1}\\
Visual Computing Group, Ping An Property \& Casualty Insurance\\
\texttt{o0o@o0oo0o.cc, tangxint@gmail.com, qixianbiao@gmail.com}\\
\AND
Chun-Guang Li \\
School of Artificial Intelligence, Beijing University of Posts and Telecommunications \\
\texttt{lichunguang@bupt.edu.cn} \\
\AND
Rong Xiao \\
Visual Computing Group, Ping An Property \& Casualty Insurance\\
\texttt{xiaorong@gmail.com} 
}

\iclrfinalcopy 
\begin{document}
\setlength{\abovedisplayskip}{0pt}
\setlength{\belowdisplayskip}{0pt}
\setlength{\abovedisplayshortskip}{0pt}
\setlength{\belowdisplayshortskip}{0pt}

\maketitle
\begin{abstract}
Graph Neural Networks (GNNs) have attracted considerable attention and have emerged as a new promising paradigm to process graph-structured data. 
GNNs are usually stacked to multiple layers and the node representations in each layer are computed through propagating and aggregating the neighboring node features with respect to the graph. 
By stacking to multiple layers, GNNs are able to capture the long-range dependencies among the data on the graph and thus bring performance improvements.  
To train a GNN with multiple layers effectively, some normalization techniques (e.g., node-wise normalization, batch-wise normalization) are necessary.  
However, the normalization techniques for GNNs are highly task-relevant and different application tasks prefer to different normalization techniques, which is hard to know in advance. 
To tackle this deficiency, in this paper, we propose to learn graph normalization by optimizing a weighted combination of normalization techniques at four different levels, including node-wise normalization, adjacency-wise normalization, graph-wise normalization, and batch-wise normalization, 
in which the adjacency-wise normalization and the graph-wise normalization are newly proposed in this paper to take into account the local structure and the global structure on the graph, respectively.
By learning the optimal weights, we are able to automatically select a single best or a best combination of multiple normalizations for a specific task. 
We conduct extensive experiments on benchmark datasets for different tasks, including node classification, link prediction, graph classification and
graph regression, and confirm that the learned graph normalization leads to competitive results and that the learned weights suggest the appropriate normalization techniques for the specific task. Source code is released here \href{https://github.com/cyh1112/GraphNormalization}{https://github.com/cyh1112/GraphNormalization}.

\end{abstract}

\section{Introduction}
\label{sec_introduction}
Graph Neural Networks (GNNs) have shown great popularity due to their efﬁciency in learning on graphs for various application areas, such as natural language processing \citep{yao2019graph, liu2019graph, zhang2018graph},  computer vision \citep{wang2018zero, li2020spatial, cheng2020skeleton}, point cloud \citep{shi2020point, liu2019relation}, drug discovery \citep{lim2019predicting}, citation networks \citep{DBLP:journals/corr/KipfW16}, social networks \citep{chen2018fastgcn} and recommendation \citep{fan2019graph, wu2019session}. A graph consists of nodes and edges, where nodes represent individual objects and edges represent relationships among those objects. In the GNN framework, the node or edge representations are 
alternately updated by propagating information along the edges of a graph via non-linear transformation and aggregation functions \citep{wu2020comprehensive, zhang2018graph}. GNN captures long-range node dependencies via stacking multiple message-passing layers, allowing the information to propagate for multiple-hops \citep{xu2018representation}. Then, the node or edge representations can be used for downstream tasks such as node classification, link prediction, and graph regression and classification.  

In essence, GNN is a new kind of neural networks which exploits neural network operations over graph structure. Among the numerous kinds of GNNs \citep{Bruna2014SpectralNA,Defferrard2016ConvolutionalNN,Chen2019OnTE,Maron2019ProvablyPG,Xu2019HowPA}, message-passing GNNs \citep{scarselli2009graph, Li2016GatedGS, DBLP:journals/corr/KipfW16, Velickovic2018GraphAN, DBLP:journals/corr/abs-1711-07553} have been the most widely 
used due to their ability to leverage the basic building blocks of deep learning such as batching, normalization and residual connections. To update the feature representation of a node, \citet{DBLP:journals/corr/KipfW16} proposed a Graph ConvNets (GCN) to employ an averaging operation over the neighborhood node with the same weight value for each of its neighbors. GraphSage \citep{Hamilton2017InductiveRL} samples a fixed-size neighborhood of each node and performs mean aggregator or LSTM-based aggregator over the neighbors. In Graph Attention Networks (GAT) \citep{Velickovic2018GraphAN}, an attention mechanism is incorporated into the propagation step. It updates the feature representation of each code via a weighted sum of adjacent node representations. \citet{Monti2017GeometricDL} has present MoNet which designs a Gaussian kernel with learnable parameters to assign different weights to neighbors. GatedGCN \citep{DBLP:journals/corr/abs-1711-07553} has achieved state-of-art results on many datasets \citep{dwivedi2020benchmarking}. GatedGCN explicitly introduces edge features at each layer and updates edge features by considering the feature representations of these two connected nodes of the edge. GatedGCN designs edge gates to assign different weights to different elements of each neighboring node representation. In addition, GatecGCN uses residual shortcut, batch normalization, and activation function to update the node representations. More details about GNNs are provided in Appendix \ref{appendix:gnn}. 

\begin{figure}[!t]
	\centering
    \includegraphics[height=5cm]{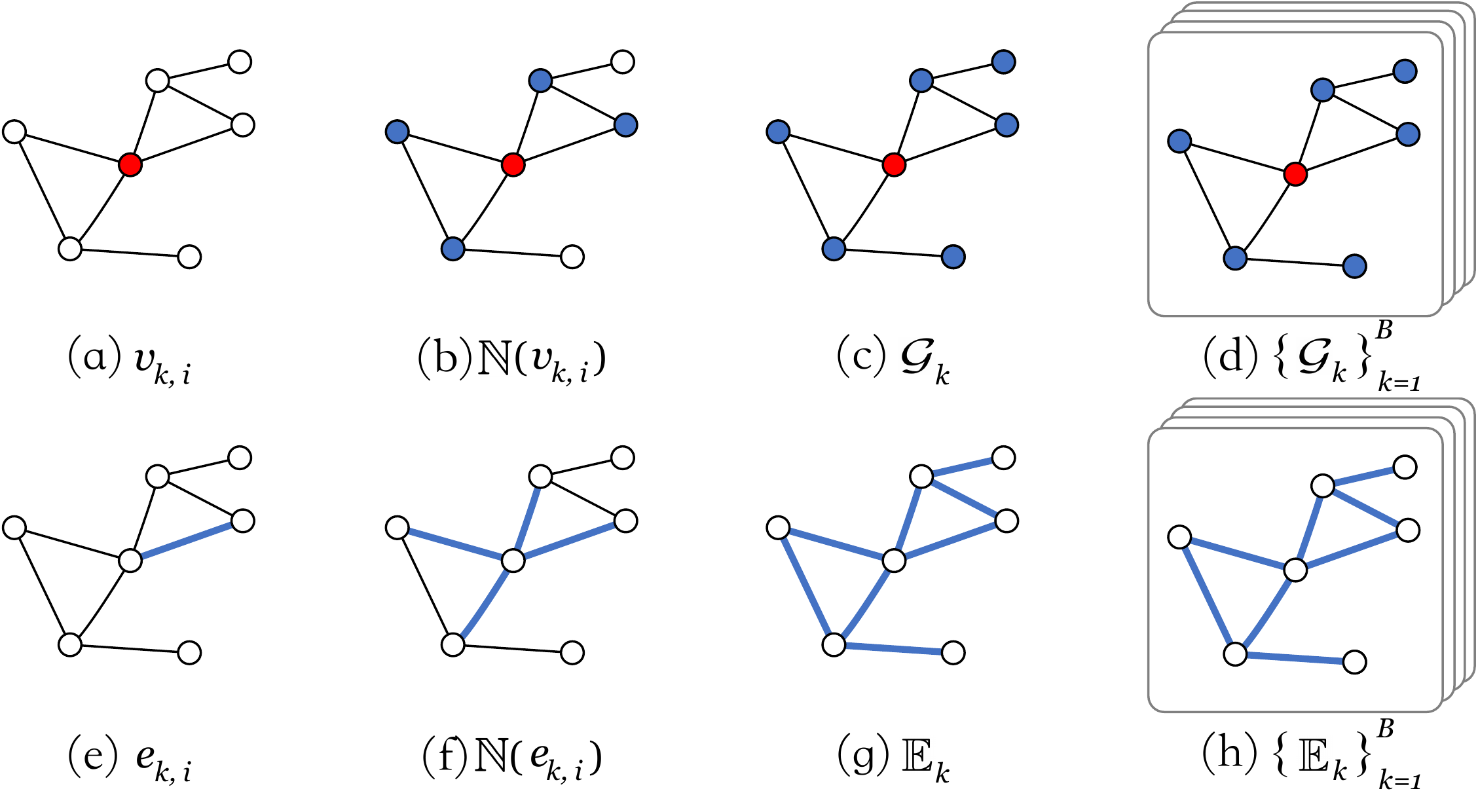}
	\caption{Illustration for four normalization methods on graph. The input node representations are normalized on four levels. (a) Node-wise. (b) Adjacency-wise. (c) Graph-wise. and (d) Batch-wise. We also extend these four normalization methods to edge features as shown in (e), (f), (g), and (h). }
   \label{graphnorm_figure}
\end{figure}

It is well accepted that normalization techniques \citep{pmlr-v37-ioffe15, ba2016layer, Wu2018GroupN, Ulyanov2016InstanceNT} are the critical ingredients to effectively train deep neural networks. 
Batch normalization (BN) \citep{pmlr-v37-ioffe15} 
is widely used in training deep neural networks 
\citep{He2016DeepRL,Huang2017DenselyCC,silver2017mastering} to 
perform global normalization along the batch dimension. Hereafter various normalization methods have been developed from different perspectives. For instance, layer normalization (LN) \citep{ba2016layer} and group normalization \citep{Wu2018GroupN} operate along the channel dimension, and instance normalization \citep{Ulyanov2016InstanceNT} performs a BN-like normalization but only for a sample. 
A detailed introduction of BN and LN are provide in the 
Appendix \ref{appendix:normalization}. In addition, Switchable Normalization \citep{luo2019switchable} 
utilizes three distinct scopes to compute statistics (means and variances) including a channel, a layer, and a minibatch. In graph neural networks, \citep{dwivedi2020benchmarking} utilizes BN for each graph propagation layer to train GNNs. \citet{Zhao2020PairNormTO} introduces a novel normalization layer denoted as PAIRNORM to mitigate the over-smoothing problem, which prevents all node representations from homogenization. PAIRNORM focuses on differentiating the distances between different node pairs. Unfortunately, it ignore the local neighbor structure and global graph structure. Each normalization method has its advantages and is suitable for some particular tasks. For instance, BN has achieved perfect performance in computer vision, however, LN outperforms BN in natural language processing \citep{DBLP:journals/corr/VaswaniSPUJGKP17}. 
Moreover, in previous work, only one of the mentioned normalization methods is selected and it is used in all normalization layers of a neural network. This will limit the performance of a neural network and it is hard to decide which normalization method is suitable for a specific task.

In this paper, we 
investigate four types of graph normalization methods, including node-wise normalization, adjacency-wise normalization, graph-wise normalization and batch-wise normalization, as 
illustrated in Figure \ref{graphnorm_figure} (a)-(d). The node-wise normalization only considers its own features and compute the statistics over the elements of its feature vector. Node-wise normalization is equivalent to layer normalization in nature. According to the adjacent structure, we normalize the node feature only using the mean and variance of its adjacent neighborhoods. We terms this normalization method as adjacency-wise normalization, as shown in Figure \ref{graphnorm_figure} (b).  We also introduce a graph-wise normalization which normalizes the nodes using the mean and variance computed from each graph alone as shown in Figure \ref{graphnorm_figure} (c). BN computes the statistics (mean and variance) over a mini-batch and is named as batch-wise normalization (Figure \ref{graphnorm_figure} (d)) in this paper. Moreover, we also extend these normalization methods to handle the edge features as shown in Figure \ref{graphnorm_figure} (e)-(h). 
Then, we propose to learn graph normalization by optimizing a weighted combination of normalization techniques at four different levels, including node-wise normalization, adjacency-wise normalization, graph-wise normalization, and batch-wise normalization. By learning the optimal weights, we are able to automatically select the most suitable normalization or a mixture of multiple normalization methods for a specific task. 

The contributions of the paper can be highlighted as follows.
\begin{itemize}[leftmargin=*]
\item According to different statistical levels of mean and variance of graph, we formulate the graph normalization methods into four levels: node-wise, adjacency-wise, graph-wise, and batch-wise. 
To the best of our knowledge, the adjacency-wise normalization and graph-wise normalization are proposed for training GNNs for the first time.

\item We propose a novel framework to learn graph normalization by optimizing a weighted combination of normalization methods at different levels. By learning the optimal weights, we are able to automatically select the best normalization method or the best combination of multiple normalization methods for a specific task. We observe that graph-wise normalization performs well on some node classification tasks and batch-wise normalization yields better performance on graph classification and regression tasks, which is consistent with the distribution of the learned dominant weights in our proposal. 

\item We conduct extensive experiments on several benchmark datasets with different tasks to 
quantitatively evaluate the listed four normalization methods and the learned normalization methods in various scenarios.

\end{itemize}

\section{Graph Normalization}
Suppose that we have $N$ graphs $\mathcal{G}_1, \mathcal{G}_2,..., \mathcal{G}_N$ in a mini-batch. Let $\mathcal{G}_k=(\mathbb{V}_k, \mathbb{E}_k)$ be the $k$-th graph, where $\mathbb{V}_k$ is the set of nodes and $\mathbb{E}_k$ is the set of edges. GNNs propagate information along the edges $\mathbb{E}$ of a graph to update the representation of each node for downstream tasks, such as node classification, link prediction and graph classification. We denote $v_{k,i}$ as the $i$-th node of the $\mathcal{G}_k$ and $h_{v_{k,i}}\in\mathbb{R}^{d}$ as the corresponding node feature. $h_{v_{k,i}}^j$ represents the $j$th element of node feature $h_{v_{k,i}}$. $\mathbb{N}(v_{k,i})$ represents the neighbors of node $v_{k,i}$ and $v_{k,i}$. 
\subsection{Normalization Methods from Different Levels} 
Currently, most GNNs 
use BN, which computes statistics over a mini-batch, after the message-passing layer to normalize feature representation. 
Note that graph data contains rich structural information. In this paper, by considering the structure information at different scales, we present and rewrite several normalization methods for graph-structure data. 

{\bf{Node-wise Normalization.}} For the node $v_{k,i}$, its feature $h_{v_{k,i}}$ has $d$ elements, as Figure \ref{graphnorm_figure} (a). Each of them may have different distribution and we can normalize them to reduce the "covariate shift" problem. Then, according to layer normalization, $h_{v_{k,i}}$ can be normalized by itself:
\begin{equation}
\label{normalization_feature_eqn}
\footnotesize
\begin{aligned}
& \hat{h}_{v_{k,i}}^{(n)} =\frac{h_{v_{k,i}}-\mu^{(n)}_{k,i}\bm{1}}{\sigma_{k,i}^{(n)}} \\
& \mu^{(n)}_{k,i}=\frac{1}{d}\sum_{j=1}^{d}h_{v_{k,i}}^j,\quad \sigma_{k,i}^{(n)} =\sqrt{\frac{1}{d}\sum_{j=1}^{d}(h_{v_{k,i}}^j-\mu^{(n)}_{k,i})^2},
\end{aligned}
\end{equation}
where $\mu^{(n)}_{k,i}$ and $\sigma_{k,i}^{(n)}$ are the mean and variance along the feature dimension for node $v_{k,i}$, and $\bm{1}\in\mathcal{R}^{d}$ represents a $d$-dimension vector filled with the scale value $1$. If we normalize the edge features, each edge feature (Figure \ref{graphnorm_figure} (e)) is processed individually as Equation (\ref{normalization_feature_eqn}). Essentially, node-wise normalization is equivalent to applying LN to each node of a graph.

{\bf{Adjacent-wise Normalization.}} Each node in a graph has its own neighbors. However, node-wise normalization performs normalization on each node alone and ignores the local structure of each node. For a node $v_{k,i}$ in a graph $\mathcal{G}_k$, we consider its topological structure of its adjacent nodes $\mathbb{N}(v_{k,i})$, as Figure \ref{graphnorm_figure} (b). The number of adjacent nodes $|\mathbb{N}(v_{k,i})|$ varies with the node $v_{k,i}$. $\mathbb{N}(v_{k,i})$ may only contain itself. Thus, the statistics (mean and variance) computed over $\mathbb{N}(v_{k,i})$ are unstable. Like LN, we can compute mean and variance across over all elements of feature vectors corresponding adjacent nodes. So, the adjacency-wise normalization for node $v_{k,i}$ is written as:
\begin{equation}
\label{normalization_adjacency_eqn}
\footnotesize
\begin{split}
& \hat{h}_{v_{k,i}}^{(a)} =\frac{h_{v_{k,i}}-\mu^{(a)}_{k,i}\mathbf{1}}{\sigma_{k,i}^{(a)}} \\
& \mu^{(a)}_{k,i}  =\frac{1}{|\mathbb{N}(v_{k,i})|\times d}\sum_{j'\in\mathbb{N}(v_{k,i})}\sum_{j=1}^{d}h_{v_{k,j'}}^{j}, \\
& \sigma_{k,i}^{(a)} = \sqrt{\frac{1}{|\mathbb{N}(v_{k,i})|\times d}\sum_{j'\in\mathbb{N}(v_{k,i})}\sum_{j=1}^{d}(h_{v_{k,j'}}^j-\mu^{(a)}_{k,i})^2},
\end{split}
\end{equation}
where $\mu_{k,i}^{(a)}$ and $\sigma_{k,i}^{(a)}$ are two scalars. For the edge $e_{k,i}$, as Figure \ref{graphnorm_figure} (f), the adjacent edges $\mathbb{N}(e_{k,i})$ are considered. 

{\bf{Graph-wise Normalization.}} Thirdly, nodes belonging to  $\mathcal{G}_k$ naturally are composed of a group. In order to preserve the global structure of a graph, the feature representation of each node is normalized based on the statistics computed over $\mathcal{G}_k$. So, we define graph-wise normalization for node $v_{k,i}$ as the following Equation (\ref{normalization_graph_eqn}):
\begin{equation}
\label{normalization_graph_eqn}
\footnotesize
\begin{split}
& \hat{h}_{v_{k,i}}^{(g)} =\frac{h_{v_{k,i}}-\mu^{(g)}_{k}}{\sigma_{k}^{(g)}} \\
& \mu^{(g)}_{k}=\mathop{\mathbb{E}}_{v\sim \mathcal{G}_{k}}{h_{v}}=\frac{1}{|\mathcal{G}_k|}\sum_{v_{k,i}\in \mathcal{G}_k}h_{v_{k,i}},\quad \\
& \sigma_{k}^{(g)} =\sqrt{\mathop{\mathbb{E}}_{v\sim \mathcal{G}_{k}}{(h_{v}}-\mu^{(g)}_{k})^2}=\sqrt{\frac{1}{|\mathcal{G}_k|}\sum_{v_{k,i}\in\mathcal{G}_k}(h_{v_{k,i}}-\mu^{(g)}_{k})^2}, \\
\end{split}
\end{equation} 
$\mu^{(g)}_{k}$ and $ \sigma_{k}^{(g)}$ in Equation (\ref{normalization_graph_eqn}) are the mean and standard deviation vectors of $\mathcal{G}_k$. Similarly, for the edges $\mathbb{E}_k$ of Graph $\mathcal{G}_k$ (Figure \ref{graphnorm_figure} (g)), we compute the mean and variance over all edges of $\mathcal{G}_k$.
When the task has only one graph, graph-wise normalization is similar to BN while the later uses a smoothing average updater but graph-wise normalization does not. 

{\bf{Batch-wise Normlaization.}} To keep training stable, BN is one of the most critical components. For a mini-batch, there are $N$ graphs. We can compute the mean and standard deviation across over the graphs of a mini-batch, then a node feature $h_{v_{k,i}}$ is normalized by the following equation, which is batch-wise normalization (GN\textsubscript{b}):
\begin{equation}
\label{normalization_batch_eqn}
\footnotesize
\begin{split}
& \hat{h}_{v_{k,i}}^{(b)} =\frac{h_{v_{k,i}}-\mu^{(b)}}{\sigma^{(b)}} \\
& \mu^{(b)}  =\operatorname{mean}\{h_{v_{k,i}}|\{\{v_{k,i}\}_{i=1}^{\mathcal{G}_k}\}_{k=1}^{N}\}=\frac{1}{\mathrm{n}}\sum_{k=1}^{N}\sum_{i=1}^{|\mathcal{G}_k|}h_{v_{k,i}},\\
& \sigma^{(b)}  =\operatorname{std}\{h_{v_{k,i}}|\{\{v_{k,i}\}_{i=1}^{\mathcal{G}_k}\}_{k=1}^{N}\}=\sqrt{\frac{1}{\mathrm{n}}\sum_{k=1}^{N}\sum_{i=1}^{|\mathcal{G}_k|}(h_{v_{k,i}}-\mu^{(b)})^2},
\end{split}
\end{equation}
where $\mathrm{n}$ means the total nodes of all $N$ graphs. GN\textsubscript{b} performs normalization over nodes of all $N$ graphs in a mini-batch and is the same as BN \citep{pmlr-v37-ioffe15} in nature.  

The properties of four normalization methods are summarized as follows.
\begin{itemize}[leftmargin=*]
  \item Node-wise normalization is equivalent as LN in operation. Node-wise normalization only considers each node's feature alone, but ignores the adjacent and whole graph structures.
  \item Adjacent-wise normalization takes the adjacent nodes into account. It reflects the difference between the node and its neighbors.
  \item Graph-wise normalization takes the features of all nodes in a graph into account. It embodies the difference of all the nodes in the whole graph. In ideal situation, this method can stand out the target node from the graph.
  \item Batch-wise normalization is the same as the standard batch normalization in nature. It expresses differences among the graphs. When the task only has one graph, then the batch-wise normalization is similar with the graph normalization except that momentum average is used in batch-wise but not in the graph-wise.
\end{itemize}


\subsection{Learning An United Graph Normalization}
Although we have 
defined several normalization methods for the graph-structured data, 
different tasks prefer to different normalization methods and for a specific task, it is hard to decide which normalization method should be used. Moreover, one normalization approach is utilized in all normalization layers of a GNN. This will sacrify 
\begin{equation}
\label{formulation_equation}
\scriptsize
\begin{split}
\hat{h}_{v_{k,i}}=\gamma (\lambda_{n}\odot\hat{h}_{v_{k,i}}^{(n)}+\lambda_{a}\odot\hat{h}_{v_{k,i}}^{(a)}+\lambda_{g}\odot\hat{h}_{v_{k,i}}^{(g)}+ \lambda_{b}\odot\hat{h}_{v_{k,i}}^{(b)})+\beta,
\end{split}
\end{equation}

where $\lambda_{n},\lambda_{a},\lambda_{g},\lambda_{b}\in\mathbb{R}^{d}$ are defined as the trainable gate parameters with the same dimension as $h_{v_{k,i}}$, $\gamma$ and $\beta$ denote the trainable scale and shift, respectively. $\lambda_{n},\lambda_{a},\lambda_{g},\lambda_{b}$ indicate the contribution of the corresponding normalized feature to $\hat{h}_{v_{k,i}}$. Thus, we constrain the elements of $\lambda_{n},\lambda_{a},\lambda_{g},\lambda_{b}$ to be in the range $[0,1]$ by the following scheme:
\begin{equation}
\left\{
\begin{aligned}
& \lambda_u^j = \operatorname{clip}_{[0, +\infty)}(\lambda_u^j)
 \quad j\in \{1,2,...,d\}\quad and\quad u\in \{n, a, g,b\}  \\
& 	\lambda_{u}^j =\frac{\lambda_{u}^j}{\sum_{k\in \{n, a, g, b\}}\lambda_{k}^j}, \quad j\in \{1,2,...,d\}\quad and\quad u\in \{n, a, g, b\} \\
\end{aligned}
\right.
\end{equation}
 $\{\lambda_u\}_{u\in\{n,a,g,b\}}$ in Equation (\ref{formulation_equation}) are important weights used to combine the normalized feature of each normalizer and they satisfy the summation constraint, $\sum_{u\in\{n,a,g,b\}}{\lambda_u^j}=1, j\in\{1,2,...,d\}$. Thus, in this mechanism, if a normalizer is better for a specific task, its corresponding weights may be higher than others. In GN, several normalization methods cooperate and compete with each other to improve the performance of GNNs.    

Different normalization methods are suitable for different tasks. In GN, the weights $\lambda_{n},\lambda_{a},\lambda_{g},\lambda_{b}$ are updated adaptively for a specific task. Usually, the best-performing normalization will have largest weight in the learnt weight distribution. Thus, GN can be an effective tactic to select one type of normalization method or a best combination of several normalization approaches for one specific task.

\section{Experiments}
\label{sec_experiments}
As \citet{dwivedi2020benchmarking}, we evaluate GN\textsubscript{n}, GN\textsubscript{a}, GN\textsubscript{g}, GN\textsubscript{b}, and the united Graph Normlization (GN) under three frameworks including Graph Convolution Network (GCN) \citep{DBLP:journals/corr/KipfW16}, Graph Attention Network (GAT) \citep{Velickovic2018GraphAN} and GatedGCN \citep{DBLP:journals/corr/abs-1711-07553}. We also assess the performance of GNNs without normalization layer named as ``No Norm''. We use the same experimental setup as \citet{dwivedi2020benchmarking}. 
The implementation of GCN, GAT and GatedGCN are found in GNN benchmarking framwork \footnote{https://github.com/graphdeeplearning/benchmarking-gnns}, which uses Deep Graph Library \citep{wang2019dgl} and PyTorch \citep{paszke2019pytorch}.

Experimental datasets consist of three types of tasks including node classification, link prediction, and graph classification/regression. We use all seven datasets from \citep{dwivedi2020benchmarking}, which are PATTERN, CLUSTER, SROIE, TSP, COLLAB, MNIST, CIFAR10, and ZINC.
In addition, we apply GatedGCN for key information extraction problem and evaluate the affect of different normalization methods on SROIE \citep{huang2019icdar2019}, which is used for extracting key information from receipt in ICDAR 2019 Challenge (task 3). For more information about the detailed statistics of the datasets refer to Appendix \ref{appendix:statistics}.    

The hyper-parameters and optimizers of the models and the details of the experimental settings are the same as GNN bechmarking framework \citet{dwivedi2020benchmarking}. To evaluate the performance of GNNs with different depths, we run experiments on CLUSTER and PATTERN datasets with depth \{4, 16\} respectively. For the other datasets, we fix the number of GCN layer to 4.

\subsection{Node Classification}
For node classification, its goal is to assign each node $v\in V$ to a number of classes. Hence, GNNs predict the label of each node by passing the node representation vector to a MLP. We evaluate the performance of different normalization approaches on CLUSTER and PATTERN datasets. Moreover, we apply node classification to key information extraction. SROIE consists of $626$ receipts for training and $347$ receipts for testing. Each image in the SROIE dataset is annotated with text bounding boxes (bbox) and the transcript of each text bbox. There are four entities to extract (Company, Date, Address and Total) from a receipt, as shown in Appendix \ref{appendix:sroie}. 
\subsubsection{CLUSTER and PATTERN Datasets}
\label{results_cluster}

For CLUSTER and PATTREN datasets, the average node-level accuracy weighted with respect to the class sizes is used to evaluate the performance of all models. For each model, we conduct 4 runs with different seeds \{41, 95, 35, 12\} and compute the average test accuracy as shown in Table \ref{tab:node_classification}.

Graph-wise normalization (GN\textsubscript{g}) outperforms batch-wise normalization (GN\textsubscript{b}) obviousluy in most situations. For instance, when the depth of GNNs is 4, GatedGCN with GN\textsubscript{g} achieves $9\%$ improvement over GN\textsubscript{b} on CLUSTER. Batch-wise normalization computes the statistics over a batch data and ignores the differences between different graphs. Different from GN\textsubscript{b}, GN\textsubscript{g} performs normalization only for each graph. Thus, GN\textsubscript{g} can learn the 
dedicated information of each graph and normalize the feature of each graph to a reasonable range. As we known, the performance of the adjacency-wise normalization (GN\textsubscript{a}) is similar with that of the node-wise normalization (GN\textsubscript{n}). Compared with GN\textsubscript{n}, GN\textsubscript{a} consider the neighbors of each node and gets higher accuracies. We can see that GN gets comparable results for different GNNs. The results of GN are close to the best results in most scenarios due to its flexibility and adaptability. GN adaptively learns weight combination of those normalization approaches which better adapt itself to the node classification task.  

\begin{table}[]
\centering
\resizebox{\textwidth}{!}{%
\begin{tabular}{@{}|c|c||cc|cc|cc||cc|cc|cc|@{}}
\hline
\multicolumn{2}{|l||}{Dataset} & \multicolumn{6}{l||}{\centering CLUSTER}                                                                & \multicolumn{6}{l|}{PATTERN}                                                                \\ [5pt] \hline
\multicolumn{2}{|l||}{\multirow{2}{*}{Network}} & \multicolumn{2}{l|}{GCN}     & \multicolumn{2}{l|}{GAT}     & \multicolumn{2}{l||}{GatedGCN} & \multicolumn{2}{l|}{GCN}     & \multicolumn{2}{l|}{GAT}     & \multicolumn{2}{l|}{GatedGCN} \\ [3pt]
\multicolumn{2}{|l||}{}             & Train (Acc)                 & Test (Acc) & Train (Acc) & Test (Acc) & Train (Acc)                  & Test (Acc) & Train (Acc) & Test (Acc) & Train (Acc)                 & Test (Acc) & Train (Acc)  & Test (Acc) \\ [3pt] \hline \hline
\multirow{5}{*}{\rotatebox{90}{Layer=4}} 
& No Norm  & \multicolumn{1}{l|}{54.3$\pm$1.9} & \multicolumn{1}{l|}{54.2$\pm$1.9}    & \multicolumn{1}{l|}{59.7$\pm$0.4} & \multicolumn{1}{l|}{59.0$\pm$0.3}   & \multicolumn{1}{l|}{58.0$\pm$2.8} & \multicolumn{1}{l||}{57.4$\pm$2.6} & \multicolumn{1}{l|}{61.9$\pm$0.2} & \multicolumn{1}{l|}{61.4$\pm$0.1} & \multicolumn{1}{l|}{81.8$\pm$0.7} & \multicolumn{1}{l|}{81.0$\pm$0.8} & \multicolumn{1}{l|}{82.5$\pm$3.2} & \multicolumn{1}{l|}{82.5$\pm$3.4}      \\ [3pt] \cline{2-14} 
& GN\textsubscript{n} & \multicolumn{1}{l|}{57.2$\pm$0.1} & \multicolumn{1}{l|}{57.0$\pm$0.1} & \multicolumn{1}{l|}{59.6$\pm$0.2} & \multicolumn{1}{l|}{59.0$\pm$0.2}           & \multicolumn{1}{l|}{61.1$\pm$0.9} & \multicolumn{1}{l||}{60.5$\pm$0.7} & \multicolumn{1}{l|}{64.9$\pm$0.1} & \multicolumn{1}{l|}{64.0$\pm$0.1}  & \multicolumn{1}{l|}{79.5$\pm$0.3} & \multicolumn{1}{l|}{78.7$\pm$0.4}           & \multicolumn{1}{l|}{82.7$\pm$2.7} & \multicolumn{1}{l|}{82.6$\pm$2.8}      \\ [3pt] \cline{2-14} 
& GN\textsubscript{a}  & \multicolumn{1}{l|}{58.9$\pm$0.6} & \multicolumn{1}{l|}{58.7$\pm$0.6}    & \multicolumn{1}{l|}{68.5$\pm$0.5} & \multicolumn{1}{l|}{67.9$\pm$0.5}   & \multicolumn{1}{l|}{63.5$\pm$0.9} & \multicolumn{1}{l||}{63.0$\pm$0.9} & \multicolumn{1}{l|}{66.5$\pm$1.4} & \multicolumn{1}{l|}{65.3$\pm$1.2} & \multicolumn{1}{l|}{82.0$\pm$0.4} & \multicolumn{1}{l|}{\color{red}{81.1$\pm$0.4}} & \multicolumn{1}{l|}{84.3$\pm$0.0} & \multicolumn{1}{l|}{84.5$\pm$0.0}      \\ [3pt] \cline{2-14} 
& GN\textsubscript{g}   & \multicolumn{1}{l|}{68.7$\pm$0.3} & \multicolumn{1}{l|}{\color{violet}{67.0$\pm$0.1}} & \multicolumn{1}{l|}{69.5$\pm$0.1} & \multicolumn{1}{l|}{\color{violet}{68.1$\pm$0.1}} & \multicolumn{1}{l|}{70.6$\pm$0.1} & \multicolumn{1}{l||}{\color{red}{69.3$\pm$0.0}} & \multicolumn{1}{l|}{80.2$\pm$0.1} & \multicolumn{1}{l|}{\color{red}{77.3$\pm$0.1}} & \multicolumn{1}{l|}{83.6$\pm$0.1} & \multicolumn{1}{l|}{\color{violet}{79.2$\pm$0.2}} & \multicolumn{1}{l|}{85.1$\pm$0.0} & \multicolumn{1}{l|}{\color{violet}{85.1$\pm$0.0}}      \\ [3pt] \cline{2-14} 
& GN\textsubscript{b}  &\multicolumn{1}{l|}{55.1$\pm$1.8} & \multicolumn{1}{l|}{54.3$\pm$1.5} & \multicolumn{1}{l|}{59.3$\pm$0.4} & \multicolumn{1}{l|}{58.6$\pm$0.3}  & \multicolumn{1}{l|}{61.4$\pm$0.2} & \multicolumn{1}{l||}{60.3$\pm$0.1}  &\multicolumn{1}{l|}{64.8$\pm$0.2} & \multicolumn{1}{l|}{63.8$\pm$0.1}  & \multicolumn{1}{l|}{78.3$\pm$1.2} & \multicolumn{1}{l|}{76.3$\pm$1.0}  & \multicolumn{1}{l|}{84.5$\pm$0.1} & \multicolumn{1}{l|}{84.5$\pm$0.1}      \\ [3pt] \cline{2-14} 
& GN   & \multicolumn{1}{l|}{68.3$\pm$0.3} &\multicolumn{1}{l|}{\color{red}{67.4$\pm$0.2}} 
& \multicolumn{1}{l|}{68.9$\pm$0.2}  &\multicolumn{1}{l|}{\color{red}{68.2$\pm$0.2}} 
& \multicolumn{1}{l|}{69.8$\pm$0.3}  &\multicolumn{1}{l||}{\color{violet}{69.3$\pm$0.1}}
& \multicolumn{1}{l|}{79.0$\pm$0.4}  &\multicolumn{1}{l|}{\color{violet}{76.5$\pm$0.4}}
& \multicolumn{1}{l|}{$81.7\pm$0.7}  &\multicolumn{1}{l|}{\color{violet}{$79.2\pm$0.7}} 
& \multicolumn{1}{l|}{85.3$\pm$0.3}  &\multicolumn{1}{l|}{{\color{red}85.2$\pm$0.2}} \\ [3pt] \hline \hline
\multirow{5}{*}{\rotatebox{90}{Layer=16}}             
& No Norm  & \multicolumn{1}{l|}{85.3$\pm$0.5} & \multicolumn{1}{l|}{72.4$\pm$0.1}    & \multicolumn{1}{l|}{63.6$\pm$2.5} & \multicolumn{1}{l|}{63.4$\pm$2.3}   & \multicolumn{1}{l|}{80.6$\pm$1.3} & \multicolumn{1}{l||}{71.2$\pm$0.4} & \multicolumn{1}{l|}{82.8$\pm$0.4} & \multicolumn{1}{l|}{\color{red}{82.9$\pm$0.4}} & \multicolumn{1}{l|}{73.4$\pm$0.4} & \multicolumn{1}{l|}{69.7$\pm$0.2} & \multicolumn{1}{l|}{85.6$\pm$0.0} & \multicolumn{1}{l|}{\color{violet}{85.7$\pm$0.0}}      \\ [3pt] \cline{2-14} 
& GN\textsubscript{n}   & \multicolumn{1}{l|}{63.6$\pm$7.0} & \multicolumn{1}{l|}{63.2$\pm$6.9}         & \multicolumn{1}{l|}{83.8$\pm$0.5} & \multicolumn{1}{l|}{72.2$\pm$0.4}           & \multicolumn{1}{l|}{84.6$\pm$0.8} & \multicolumn{1}{l||}{73.8$\pm$0.2}          & \multicolumn{1}{l|}{76.7$\pm$0.8} & \multicolumn{1}{l|}{71.4$\pm$0.3}           & \multicolumn{1}{l|}{87.7$\pm$0.7} & \multicolumn{1}{l|}{\color{violet}{81.8$\pm$0.5}}           & \multicolumn{1}{l|}{85.6$\pm$0.0} & \multicolumn{1}{l|}{\color{red}{85.8$\pm$0.0}}       \\ [3pt] \cline{2-14} 
& GN\textsubscript{a}   & \multicolumn{1}{l|}{66.7$\pm$1.5} & \multicolumn{1}{l|}{66.0$\pm$1.5} & \multicolumn{1}{l|}{85.6$\pm$0.6} & \multicolumn{1}{l|}{\color{violet}{73.2$\pm$0.2}}           & \multicolumn{1}{l|}{84.7$\pm$0.6} & \multicolumn{1}{l||}{\color{violet}{74.1$\pm$0.3}}           & \multicolumn{1}{l|}{74.9$\pm$1.7} & \multicolumn{1}{l|}{70.7$\pm$0.7}           & \multicolumn{1}{l|}{84.8$\pm$0.3} & \multicolumn{1}{l|}{\color{red}{82.8$\pm$0.5}}          & \multicolumn{1}{l|}{85.6$\pm$0.0} & \multicolumn{1}{l|}{\color{red}{85.8$\pm$0.0}}     \\ [3pt] \cline{2-14} 
& GN\textsubscript{g}   & \multicolumn{1}{l|}{87.2$\pm$0.4} & \multicolumn{1}{l|}{\color{violet}{72.5$\pm$0.2}}     & \multicolumn{1}{l|}{91.9$\pm$0.3} & \multicolumn{1}{l|}{\color{red}{73.4$\pm$0.1}}           & \multicolumn{1}{l|}{90.9$\pm$0.5} & \multicolumn{1}{l||}{\color{violet}{74.5$\pm$0.1}}   & \multicolumn{1}{l|}{98.9$\pm$0.1} & \multicolumn{1}{l|}{76.3$\pm$0.2}           & 
\multicolumn{1}{l|}{92.8$\pm$0.1} & \multicolumn{1}{l|}{\color{violet}{81.2$\pm$0.2}}  & \multicolumn{1}{l|}{86.7$\pm$0.2} &  \multicolumn{1}{l|}{85.3$\pm$0.1}   \\ [3pt] \cline{2-14} 
& GN\textsubscript{b} &\multicolumn{1}{l|}{67.6$\pm$3.7} & \multicolumn{1}{l|}{65.1$\pm$2.6} &\multicolumn{1}{l|}{83.9$\pm$0.6} &\multicolumn{1}{l|}{72.2$\pm$0.3}           & \multicolumn{1}{l|}{88.2$\pm$1.0}  &\multicolumn{1}{l||}{73.7$\pm$0.3} & \multicolumn{1}{l|}{79.0$\pm$1.6} & \multicolumn{1}{l|}{72.0$\pm$0.3}           & \multicolumn{1}{l|}{91.9$\pm$0.6} & \multicolumn{1}{l|}{80.2$\pm$0.2}           & \multicolumn{1}{l|}{86.1$\pm$0.2} & \multicolumn{1}{l|}{\color{violet}{85.7$\pm$0.1}}   \\ [3pt] \cline{2-14}
& GN &\multicolumn{1}{l|}{85.8$\pm$0.4} &\multicolumn{1}{l|}{\color{red}{73.8$\pm$0.2}}           & \multicolumn{1}{l|}{87.3$\pm$2.1} &\multicolumn{1}{l|}{72.6$\pm$0.6}           & \multicolumn{1}{l|}{88.6$\pm$0.4} &\multicolumn{1}{l||}{\color{red}{75.8$\pm$0.2}} & \multicolumn{1}{l|}{93.6$\pm$1.9} &\multicolumn{1}{l|}{\color{violet}{77.8$\pm$0.3}}           & \multicolumn{1}{l|}{95.6$\pm$0.5} &\multicolumn{1}{l|}{79.2$\pm$0.4}           & 
\multicolumn{1}{l|}{87.3$\pm$0.3} &\multicolumn{1}{l|}{\color{violet}{85.7$\pm$0.1}}        \\  [3pt] \hline
\end{tabular}
}
\caption{Node classification results on the CLUSTER and PATTERN. Results are averaged over four different random seeds. {\color{red}{Red}}: the best model,{\color{violet}{Violet}}: good models.}
\label{tab:node_classification}
\end{table}

\subsubsection{SROIE Dataset}

In this experiment, we test the performance of GatedGCN with several different normalization approaches for extracting key information from a receipt. For a receipt image, each text bounding box is label with five classes, which are Total, Date, Address, Company and Other. Thus, for a receipt, we treat each text bounding box as a node. Feature representation for each node will be supplied by Appendix \ref{appendix:sroie}. ``Company'' and ``Address'' usually consist of multiple text bounding boxes (nodes). If and only if all nodes of each entity are classified correctly, this entity is extracted successfully. We compute the mean accuracies for each text field and the average accuracies for each receipt as shown in Table~\ref{tab:sroie}.

\begin{table}[h]
\centering
\resizebox{0.6\textwidth}{!}{%
\begin{tabular}{|c|c|c|c|c|c|c|}
\hline
           & \multicolumn{6}{c|}{Normalization Method}    \\ \hline
Text Field & No Norm & GN\textsubscript{n} & GN\textsubscript{a} & GN\textsubscript{g} & GN\textsubscript{b} & GN \\ \hline
Total      &  87.5   & 91.9  & 74.5  &  {\bf{96.8}}     & 94.8      & 94.5  \\ \hline
Date       &  96.5   & 98.0  & 95.9  &  {\bf{98.8}}       & 97.4      & 97.4   \\ \hline
Address    & 91.6   & 92.0     &  80.0    & {\bf{94.5}}       &  93.9     & 93.6   \\ \hline
Company    & 92.2   & 93.3      &  87.8     &   94.5    &  93.0     & {\bf{94.8}}   \\ \hline
Average    & 92.0   &  94.0     &   84.6    & {\bf{96.2}}      &  94.8     & 95.1   \\ \hline
\end{tabular}%
}
\caption{Performance (accuracy) comparison of different normalization approaches.}
\label{tab:sroie}
\end{table}

We can observe that 
GN\textsubscript{g} achieves the best performance among all compared normalization methods. In the receipt, there are many nodes with only numeric texts. It is hard to differentiate the ``Total'' field from other nodes with numeric text. GN\textsubscript{g} performs well in this field and outperforms the second best by 2.0\%. We believe that the graph-wise normalization can make the 'Total' field stand out from the other bounding boxes with numeric text by aggregating the relevant anchor point information from its neighbors and removing the mean number information. Similarly, graph-wise normalization can promote extracting  information for the other three key fields. It's interesting that the graph of each receipt is special with Neighboring nodes usually belonging to different classes. Thus, the performance of adjacency-wise normalization is worse than node-wise normalization.    

\subsection{Link Prediction}
Link prediction is to predict whether there is a link between two nodes $v_i$ and $v_j$ in a graph. Two node features of $v_i$ and $v_j$, at both ends of edge $e_{i,j}$,  are concatenated to make a prediction. Then the concatenated feature is fed into a MLP for prediction. Experimental results are shown in Table~\ref{tab:link_prediction}.

All the five normalization methods achieve similar performance on TSP dataset. Compared with other normalization methods, the results of GN are very stable. For each GNN, the result of GN is comparable with the best result. In addition, GNNs without normalization layer obtains good performance.  

COLAB dataset contains only a graph with 235,868 nodes. When we use adjacency-wise normalization, we encounter out-of-memory problem. Thus, we don't report the results of GN\textsubscript{a} and GN. Compared with GNNs with normalization layer, the results of GNNs without normalization layer (No Norm) seriously decrease. GN\textsubscript{g} performs better than GN\textsubscript{b}. GatedGCN with GN\textsubscript{g} achieves the best result.     

\begin{table}[]
\footnotesize
\centering
\resizebox{\textwidth}{!}{%
\begin{tabular}{c|cc|cc|cc}
\toprule
Network                  & \multicolumn{2}{c|}{GCN} & \multicolumn{2}{c|}{GAT} & \multicolumn{2}{c}{GatedGCN} \\ 
\toprule
\multirow{2}{*}{Dataset} & \multicolumn{6}{c}{TSP} \\
& Train (F1)  & Test (F1)  & Train (F1)  & Test (F1)  & Train (F1)  & Test (F1)    \\ 
\midrule
No Norm & 0.628$\pm$0.001 & 0.627$\pm$0.001 & 0.677$\pm$0.002 & {\color{violet}{0.675$\pm$0.002}} & 0.805$\pm$0.005 &0.804$\pm$0.005 \\
 GN\textsubscript{n} & 0.635$\pm$0.001 & {\color{red}{0.634$\pm$0.001}}  & 0.663$\pm$0.008   & 0.662$\pm$0.008  & 0.810$\pm$0.003 & {\color{red}{0.808$\pm$0.003}}  \\
 GN\textsubscript{a} & 0.633$\pm$0.004 & 0.631$\pm$0.004 & 0.678$\pm$0.003 & {\color{red}{0.676$\pm$0.003}}          & 0.805$\pm$0.005           & 0.803$\pm$0.004          \\
 GN\textsubscript{g}            & 0.630$\pm$0.001           & 0.629$\pm$0.001          & 0.669$\pm$0.003           & 0.668$\pm$0.001          & 0.890$\pm$0.001           & {\color{violet}{0.806$\pm$0.001}}          \\
 GN\textsubscript{b}            & 0.633$\pm$0.001           & 0.632$\pm$0.001          & 0.673$\pm$0.004           & 0.671$\pm$0.004          & 0.791$\pm$0.002           & 0.789$\pm$0.002          \\
 GN            & 0.635$\pm$0.001           &  {\color{violet}{0.633$\pm$0.001}} & 0.673$\pm$0.001    &0.671$\pm$0.001      & 0.804$\pm$0.001  & 0.802$\pm$0.001        \\  
 \midrule
 \multirow{2}{*}{Dataset} & \multicolumn{6}{c}{COLAB} \\
& Train (Hits)  & Test (Hits)  & Train (Hits)  & Test (Hits)  & Train (Hits)     & Test (Hits)    \\ 
\midrule
 No Norm &73.02$\pm$7.03 &38.32$\pm$4.13  &64.19$\pm$4.02 &32.69$\pm$4.48         &38.55$\pm$8.13 &22.60$\pm$3.40 \\
GN\textsubscript{n} &81.81$\pm$7.40  &45.75$\pm$4.14 &95.93$\pm$0.54 & {\color{red}{51.76$\pm$0.68}} &91.72$\pm$3.40  &51.55$\pm$1.44          \\
 GN\textsubscript{g} &93.67$\pm$0.71 & {\color{red}{52.27$\pm$1.28}}   &97.11$\pm$0.65  &51.36$\pm$1.15          
 &97.50$\pm$2.52  & {\color{red}{52.71$\pm$0.36}}          \\
 GN\textsubscript{b} &91.88$\pm$0.04  & {\color{violet}{51.16$\pm$0.10}}  &97.11$\pm$0.43 & {\color{violet}{51.54$\pm$0.90}} &95.31$\pm$3.56 & {\color{violet}{51.87$\pm$0.41}}          \\
\bottomrule
\end{tabular}%
}
\caption{Link prediction results on the TSP and COLAB. {\color{red}{Red}}: the best model,{\color{violet}{Violet}}: good models.}
\label{tab:link_prediction}
\end{table}

\subsection{Graph Classification and Graph Regression}


Graph classification is to assign one label to each graph. We conduce experiments on CIFAR10 and MNIST. Average class accuracies are reported in Table~\ref{tab:graphclassification}. ZINC is a dataset for graph regression. the mean absolute error (MAE) between the predicted value and the ground truth is calculated for each group. Average MAEs also are reported in Table~\ref{tab:graphclassification}. We can see that GN\textsubscript{b} outperforms others in most cases. GN\textsubscript{g} doesn't work well on graph classification and regression. Furthermore, GN\textsubscript{g} affects the performance of GN. GN integrates the normalized features of GN\textsubscript{n}, GN\textsubscript{a}, GN\textsubscript{g}, and GN\textsubscript{b} and adptively pays more attention to GN\textsubscript{b} due to its outstanding performance. Therefore, its performance is comparable with GN\textsubscript{b}.

\begin{table}[!h]
\centering
\label{table_graphclassification_results}
\resizebox{\textwidth}{!}{%
\begin{tabular}{|c|cc|cc|cc|}
\hline
\multirow{1}{*}{Network}      & \multicolumn{2}{c|}{GCN}                               & \multicolumn{2}{c|}{GAT}                               & \multicolumn{2}{c|}{GatedGCN}                          \\ \hline
{}   & Train (Acc)     & Test (Acc)            & Train (Acc)                               & Test (Acc)     & Train (Acc)  & Test (Acc)            \\ \hline
Dataset                       & \multicolumn{6}{c|}{MNIST}                                                           \\ \hline
No Norm & \multicolumn{1}{l|}{93.62$\pm$0.72} &90.10$\pm$0.25 & \multicolumn{1}{l|}{100.00$\pm$0.00}   & 95.39$\pm$0.16 & \multicolumn{1}{l|}{100.00$\pm$0.00}   & 96.61$\pm$0.09 \\ \hline
GN\textsubscript{n} & \multicolumn{1}{l|}{96.63$\pm$0.91} & {\color{red}{90.53$\pm$0.22}} & \multicolumn{1}{l|}{100.00$\pm$0.00}   &  {\color{violet}{95.66$\pm$0.14}} & \multicolumn{1}{l|}{100.00$\pm$0.00}   & 97.23$\pm$0.12 \\ \hline
GN\textsubscript{a}                            & \multicolumn{1}{l|}{95.65$\pm$0.75} & 89.68$\pm$0.20 & \multicolumn{1}{l|}{100.00$\pm$0.00}   & 95.41$\pm$0.22 & \multicolumn{1}{l|}{100.00$\pm$0.00}   & 96.87$\pm$0.21 \\ \hline
GN\textsubscript{g}                            & \multicolumn{1}{l|}{96.90$\pm$0.52} & 86.18$\pm$0.30 & \multicolumn{1}{l|}{100.00$\pm$0.00}   & 94.74$\pm$0.13 & \multicolumn{1}{l|}{100.00$\pm$0.00}   & 96.17$\pm$0.16 \\ \hline
GN\textsubscript{b}                            & \multicolumn{1}{l|}{97.16$\pm$1.06} & {\color{violet}{90.51$\pm$0.22}} & \multicolumn{1}{l|}{99.99$\pm$0.00} & {\color{red}{95.77$\pm$0.19}} & \multicolumn{1}{l|}{100.00$\pm$0.00}   & {\color{red}{97.47$\pm$0.11}} \\ \hline
GN                            & \multicolumn{1}{l|}{97.34$\pm$0.66} &  {\color{violet}{90.46$\pm$0.17}} & \multicolumn{1}{l|}{100.00$\pm$0.00} &  {\color{violet}{95.75$\pm$0.22}} & \multicolumn{1}{l|}{100.00$\pm$0.00} &  {\color{violet}{97.41$\pm$0.17}} \\ \hline \hline
\multicolumn{1}{|l|}{Dataset} & \multicolumn{6}{c|}{CIFAR10} \\ \hline
No Norm  & \multicolumn{1}{l|}{65.87$\pm$1.68} & 54.56$\pm$0.53 & \multicolumn{1}{l|}{88.80$\pm$1.31} &62.13$\pm$0.31 & \multicolumn{1}{l|}{82.81$\pm$1.15} & 63.44$\pm$0.22 \\ \hline
GN\textsubscript{n}                            & \multicolumn{1}{l|}{73.64$\pm$1.42} & {\color{red}{55.77$\pm$0.31}} & \multicolumn{1}{l|}{87.67$\pm$0.81} & {\color{violet}{63.04$\pm$0.60}} & \multicolumn{1}{l|}{90.14$\pm$2.05} & {\color{red}{67.86$\pm$0.65}} \\ \hline
GN\textsubscript{a}                            & \multicolumn{1}{l|}{71.48$\pm$1.27} & 54.83$\pm$0.32 & \multicolumn{1}{l|}{86.80$\pm$0.70} & 62.72$\pm$0.26 & \multicolumn{1}{l|}{90.85$\pm$0.32} & 67.21$\pm$0.44 \\ \hline
GN\textsubscript{g}                            & \multicolumn{1}{l|}{71.75$\pm$2.48} & 46.41$\pm$0.29 & \multicolumn{1}{l|}{89.20$\pm$0.41} & 54.44$\pm$0.28 & \multicolumn{1}{l|}{81.29$\pm$6.37} & 52.69$\pm$3.28 \\ \hline
GN\textsubscript{b}                            & \multicolumn{1}{l|}{69.34$\pm$2.47} & {\color{violet}{55.14$\pm$0.26}} & \multicolumn{1}{l|}{89.56$\pm$1.41} & {\color{red}{64.54$\pm$0.24}} & \multicolumn{1}{l|}{95.75$\pm$0.12} & {\color{violet}{67.83$\pm$0.68}} \\ \hline
GN  & \multicolumn{1}{l|}{80.33$\pm$3.10} & 54.73$\pm$0.68 & \multicolumn{1}{l|}{94.48$\pm$1.58} & {\color{violet}{62.98$\pm$0.47}} & \multicolumn{1}{l|}{98.80$\pm$0.28} & 66.84$\pm$0.16 \\ \hline
\hline
{}   & Train (MAE)     & Test (MAE)            & Train (MAE)                               & Test (MAE)     & Train (MAE)  & Test (MAE)          \\ \hline
\multicolumn{1}{|l|}{Dataset} & \multicolumn{6}{c|}{ZINC} \\ \hline
No Norm &0.368$\pm$0.022  &0.472$\pm$0.005 &0.270$\pm$0.029 &0.490$\pm$0.001 &0.292$\pm$0.003  &0.456$\pm$0.004  \\ \hline

GN\textsubscript{n}            & \multicolumn{1}{l|}{0.349$\pm$0.019}           & {\color{red}{0.455$\pm$0.007}}          & \multicolumn{1}{l|}{0.295$\pm$0.014}           & {\color{red}{0.456$\pm$0.001}}         & \multicolumn{1}{l|}{0.260$\pm$0.021}           & {\color{red}{0.428$\pm$0.005}}          \\ \hline
GN\textsubscript{a}            & \multicolumn{1}{l|}{0.351$\pm$0.013}           & {\color{violet}{0.458$\pm$0.003}}          & \multicolumn{1}{l|}{0.291$\pm$0.013}          & {\color{violet}{0.458$\pm$0.001}}          & \multicolumn{1}{l|}{0.274$\pm$0.023}           & {\color{violet}{0.437$\pm$0.001}}          \\ \hline
GN\textsubscript{g}            &\multicolumn{1}{l|}{0.263$\pm$0.033}           & 0.547$\pm$0.029          &\multicolumn{1}{l|}{0.228$\pm$0.010}           & 0.519$\pm$0.001          &\multicolumn{1}{l|}{0.216$\pm$0.019}           & 0.507$\pm$0.003          \\ \hline
GN\textsubscript{b}            &\multicolumn{1}{l|}{0.346$\pm$0.019}           & 0.465$\pm$0.009          &\multicolumn{1}{l|}{0.308$\pm$0.028}           & 0.480$\pm$0.003          &\multicolumn{1}{l|}{0.280$\pm$0.013}           & {\color{violet}{0.431$\pm$0.007}}          \\ \hline
GN            &\multicolumn{1}{l|}{0.357$\pm$0.017}          &0.486$\pm$0.007          &\multicolumn{1}{l|}{0.298$\pm$0.018}           & 0.483$\pm$0.005          &\multicolumn{1}{l|}{0.275$\pm$0.011}           & 0.458$\pm$0.003          \\ \hline
\end{tabular}%
}
\caption{Experimental results on MNIST, CIFAR10 and ZINC. {\color{red}{Red}}: the best model,{\color{violet}{Violet}}: good models.}
\label{tab:graphclassification}
\end{table}

\subsection{Analysis}
The above experimental results indicates that GN\textsubscript{g} outperforms batch normalization on most node classification tasks. For each single normalization method, it performs very well on some datasets, while its performance may decrease sharply on other datasets. Meanwhile, our proposed GN, which integrates several normalization methods into a framework, achieves competitive results compared with the best single normalization method on various datasets. 

\begin{figure}[h]
	\centering
	\includegraphics[scale=0.40]{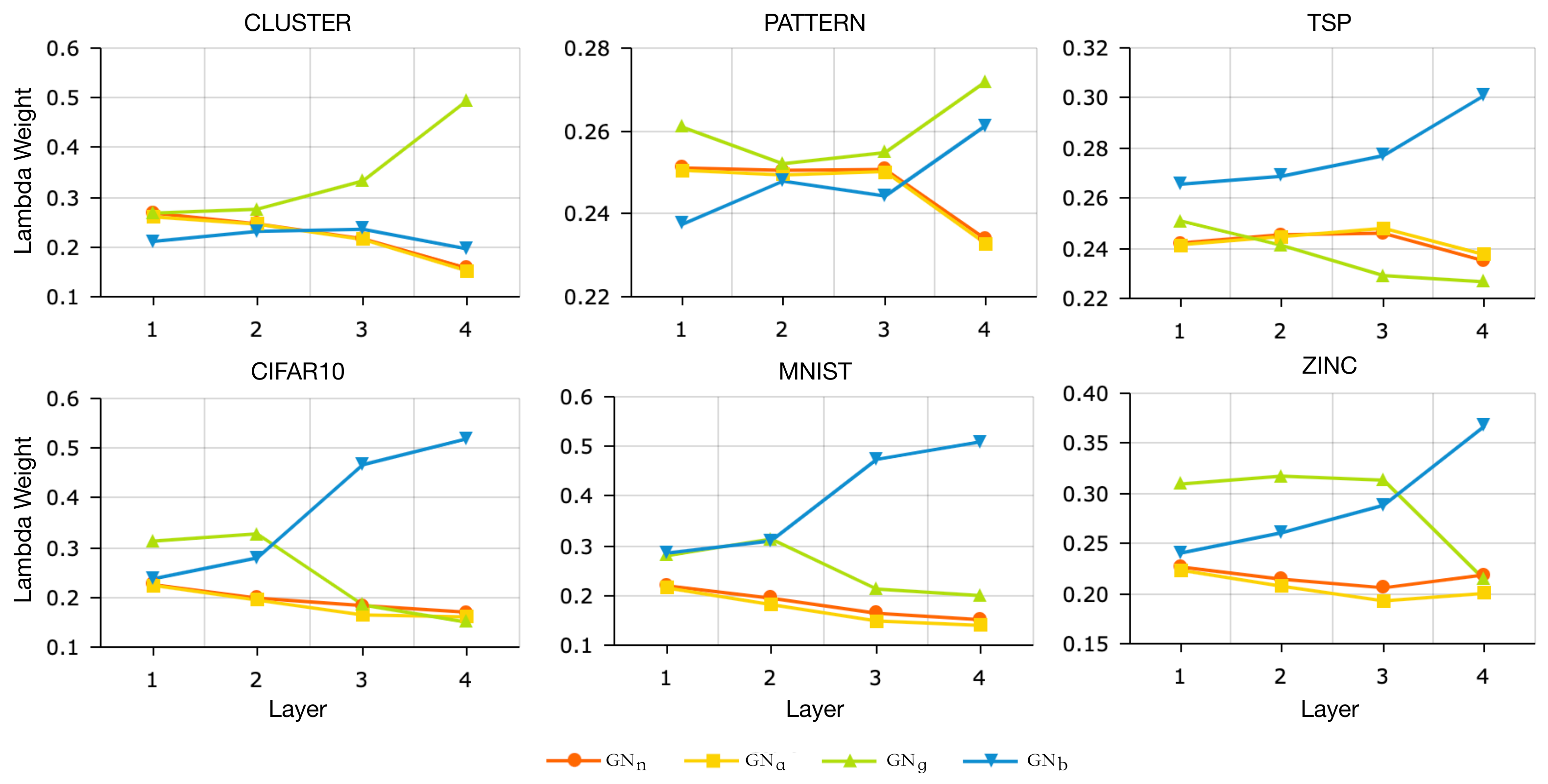}
	\caption{Learnt weight distributions of four normalization methods along with layers on different tasks.}
	\label{fig:ratios}
\end{figure}

In this part, we analyze the effect of each normalization method on different datasets. GN adaptively combines the results of several normalization methods and $\{\lambda_u\}_{u\in {n,a,g,b}}$ in Equation (\ref{formulation_equation}) indicate the importance weight vector of the corresponding normalizer, respectively. We initialize the important weights $\{\lambda_u\}_{u\in {n,a,g,b}}$ in each layer to the equal values (\textit{i.e.} 1/4). In the training phase, the values of $\{\lambda_u\}_{u\in {n,a,g,b}}$ changes between 0 and 1. We investigate the averaged weights in different layers of GatedGCN on several datasets. We get the important weights of each normalizer in each layer from the optimized model. Since $\lambda_u\in\mathbb{R}^{d}$, then the averaged weights of each normalizer is calculated over all of the $d$ elements of $\lambda_u$. According to Figure \ref{fig:ratios}, the weights of each normalizer not only are distinct for different datasets, but also vary for different layers. It implies that distinct layers may have their own preference of normalization methods to achieve good performance. It's interesting that the weights of GN\textsubscript{g} are larger than others on node classification tasks while GN\textsubscript{b} is more importance on other tasks. GN has the ability to automatically choose the suitable normalizers for a specific task. 

Furthermore, for each dataset, we select the two best normalizers and integrate them into a new normalization method like Equation (\ref{formulation_equation}). From Table \ref{tab:combine}, the combined normalizer achieves the comparable results with the best normalization method for each dataset. Therefore, these results show that the important weights indicate whether the corresponding normalization method is suitable for the current task.  

\begin{table}[]
\centering
\resizebox{0.8\textwidth}{!}{%
\begin{tabular}{l|llllllll}
\hline
        & CLUSTER (Acc) & PATTERN (Acc) & SROIE (Acc) & TSP (F1)  & MNIST (Acc) & CIFAR10 (Acc) & ZINC (MAE)\\ \hline
GN\textsubscript{n}     & $-$  & $-$ &   $-$    &  80.83   &  97.23           & 67.86      &  0.4283    \\
GN\textsubscript{a}      & 63.02 & 84.53 & $-$  & $-$   &  $-$  & $-$  & $-$      \\
GN\textsubscript{g}      &  69.31  &85.07  &96.2 & 80.61 & $-$  &$-$ & $-$     \\
GN\textsubscript{b}      & $-$ & $-$ &94.8 &$-$  &  97.47     & 67.83        &  0.4311    \\ \hline
Combine & 69.16 &84.64   &95.4  &81.11     & 97.52      & 67.88   & 0.4371     \\ \hline
\end{tabular}%
}
\caption{Performance of different normalization methods on several datasets. For each dataset, we give the performance of two best normalization methods and a new normalization method combined these two best normalizer like Equation (\ref{formulation_equation}).}
\label{tab:combine}
\end{table}

\section{Conclusion}
In this paper we formulate four graph normalization methods, node-wise, adjacent-wise, graph-wise, and graph-wise according to different statistical levels of mean and variance. Node-wise normalization only considers its own statistical information. Adjacent-wise and graph-wise takes local and global graph structures into account. BN computes the statistics in a mini-batch level. Different normalization methods perform variously in different tasks. We observe graph-wise and adjacent-wise normalizations perform well on some node classification tasks, batch-wise normalization shows better performance on graph classification and regression tasks. Therefore, we propose to learn an effective Graph Normalization (GN) by optimizing a weighted combination of node-wise normalization, adjacency-wise normalization, graph-wise normalization, and batch-wise normalization. Through analyzing the distributions of weights, we can select one or a combinations of several normalization for one specific task.

\bibliography{iclr2021_conference}
\bibliographystyle{iclr2021_conference}

\appendix
\section{Graph Neural Networks}
\label{appendix:gnn}
Graph neural networks \citep{DBLP:journals/corr/KipfW16, Velickovic2018GraphAN} are effective in learning graph representations. For node $v$, GNNs update its representation by utilizing itself and its adjacent neighbors. To capture high-order structure information of the graph,  GNNs learn a new feature representation of each node over multiple layers. In a layer of GNNs, each node $v$ sends a ``message''- its feature representation, to the nodes in $\mathcal{N}(v)$; and then the feature representation of $v$ is updated according to all collected information from the neighborhood $\mathcal{N}(v)$. Mathematically, at the $\ell$-th layer, we have,
\begin{equation}
\label{subsec_gnn_message_pass_equ}
h_{v}^{\ell+1}=\psi^{\ell+1}(\mathcal{C}\{h_{v}^{\ell}, \mathcal{M}\{\phi^{\ell+1}(h_u^{\ell})|u\in\mathcal{N}(v)\}\})
\end{equation}
where $h_u^{\ell}$ denote the feature vector at layer $\ell$ of node $u\in\mathcal{N}(v)$, $\psi$ and $\phi$ are learnable functions, $\mathcal{M}$ is the aggregation function for nodes in $\mathcal{N}(v)$, and $\mathcal{C}$ is utilized to combine the feature of node $v$ and its neighbors. Especially,  the initial node representation $h_v^{0}=x_v$ represents the original input feature vector of node $v$.

Graph ConvNets\citep{DBLP:journals/corr/KipfW16} treats each neighbor node $u$ equally to update the representation of a node $v$ as:
\begin{equation}
\label{subsec_gnn_gcn_equ}
h_{v}^{\ell+1}= \operatorname{ReLU}(\frac{1}{deg_v}\sum_{u\in\mathcal{N}(v)}{W^{\ell}h_{u}^{\ell}}),
\end{equation}
where $W\in\mathcal{R}^{d\times d}$, $deg_v$ is the in-degree of node $v$. One graph convolutional layer only considers immediate neighbors. To use neighbors within K hops, in practice, multiple GCN layers are stacked. All neighbors contribute equally in the information passing of GCN. One of key issue of the GCN is its over-smoothing problem. This issue can be partially eased by residual shortcut across layers. One effective approach is to use spatial GNNs, such as GAT \citep{Velickovic2018GraphAN} and GatedGCN \citep{DBLP:journals/corr/abs-1711-07553}.

GAT \citep{Velickovic2018GraphAN} learns to assign different weight to adjacent nodes by adopting attention mechanism. In GAT, the feature representation of $v$ can be updated by:
\begin{equation}
\label{subsec_gat_update_equ}
h_{v}^{\ell+1}= \sigma(\sum_{u\in\mathcal{N}(v)}{e_{u,v}^{\ell}W^{\ell}h_{u}^{\ell}}),
\end{equation} 
where $e_{u,v}^{\ell}$ measures the contribution of node $u$'s feature to node $v$:
\begin{equation}
\label{subsec_gat_attention_equ}
e_{u,v}^{\ell}= \frac{exp(g(\alpha^T[W^{\ell}h_u^{\ell}||W^{\ell}h_v^{\ell}]))}{\sum_{k\in\mathcal{N}(v)}exp(g(\alpha^T[W^{\ell}h_k^{\ell}||W^{\ell}h_v^{\ell}]))},
\end{equation} 
where $g(\cdot)$ is a $\operatorname{LeaklyReLU}$ activation function, $\alpha$ is a weight vector and $||$ is the concatenation operation. Similar to \citet{DBLP:journals/corr/VaswaniSPUJGKP17}, to expand GAT's expressive capability and stabilize the learning process, multi-head attention is employed in GAT. 
GAT has achieved an impressive improvement over GCN on node classification tasks. However, as the number of graph convolutional layers increases, nodes representations will converge to the same value. The over-smoothing problem still exists.

To mitigate the over-smoothing problem, GatedGCN \citep{DBLP:journals/corr/abs-1711-07553} integrates gated mechanism \citep{hochreiter1997long}, batch normalization \citep{pmlr-v37-ioffe15}, and residual connections \citep{He2016DeepRL} into the network design. Unlike GCNs which treats all edges equally, GatedGCN uses an edge gated mechanism to give different weights to different nodes. So, for node $v$, the formulation of updating the feature representation is:
\begin{equation}
\label{subsec_gnn_gatedgcn_equ}
h_{v}^{\ell+1} =h_{v}^{\ell}+ \operatorname{ReLU}(\operatorname{BN}(W^{\ell}h_{v}^{\ell}+\sum_{u\in\mathcal{N}(v)}{e_{vu}^{\ell}\odot U^{\ell}h_{u}^{\ell}})),
\end{equation} 
where $W^{\ell},U^{\ell}\in\mathcal{R}^{d\times d}$, $\odot$ is the Hadamard product, and the  edge gates $e_{vu}^{\ell}$ are defined as:
\begin{equation}
\begin{split}
& e_{v,u}^{\ell} = \frac{\sigma(\hat{e}_{v,u}^{\ell})}{\sum_{u^{\prime}\in\mathcal{N}(v)}{\sigma(\hat{e}_{v,u^{\prime}}^{\ell})}+\varepsilon},  \\
& \hat{e}_{v,u}^{\ell} = \hat{e}_{v,u}^{\ell-1}+\operatorname{ReLU}(\operatorname{BN}(A^{\ell} h_{v}^{\ell-1}+B^{\ell} h_{u}^{\ell-1}+C^{\ell} \hat{e}_{v,u}^{\ell-1})),
\end{split}
\end{equation}
where $\sigma$ is the sigmoid function, $\varepsilon$ is a small fixed constant, $A^{\ell},B^{\ell},C^{\ell}\in R^{d\times d}$. Different from traditional GNNs, GatedGCN explicitly considers edge feature $\hat{e}_{v,u}$ at each layer.

\section{Normalization methods}
\label{appendix:normalization}

\subsection{Batch Normalization}
Batch normalization \citep{pmlr-v37-ioffe15} (BN) has become one of the critical components in a deep neural network, which normalizes the features by using the statics computed within a mini-batch. BN can reduce the internal covariate shift problem and accelerate training. We briefly introduce the formulation of BN. Firstly, $H=[h_1,h_2,...,h_B]^T\in\mathcal{R}^{B\times d}$ is denoted as the input of a normalization layer, where $B$ is the batch size and $h_i$ represents a sample. Then, $\mu^{(B)}\in\mathcal{R}^{d}$ and $\sigma^{(B)}\in\mathcal{R}^{d}$ mean the mean and the variance of $H$ along the batch dimension, respectively. BN normalizes each dimension of features using  $\mu^{(B)}$ and $\sigma^{(B)}$ as:
\begin{equation}
\label{subsec_bn_equ}
\begin{split}
\hat{H} & =  \gamma\frac{H-\mu^{(B)}}{\sigma^{(B)}}+\beta, \\
\mu^{(B)} & =  \frac{1}{B}\sum_{i=1}^{B}h_i, \quad \sigma^{(B)} =\sqrt{ \frac{1}{B}\sum_{i=1}^{B}(h_i-\mu^{(B)})^2},  \\
\mu & =\alpha\mu+(1-\alpha)*\mu^{(B)}, \quad \sigma^2=\alpha\sigma^2+(1-\alpha)(\sigma_B)^2  \\
\end{split}
\end{equation}
where $\gamma$ and $\beta$ are trainable scale and shift parameters, respectively. In Equation (\ref{subsec_bn_equ}), $\mu$ and $\sigma$ denote the running mean and variance to approximate the mean and the variance of the dataset. So, during testing, they are used for normalization.  

\subsection{Layer Normalization}
\label{subsec_layer_norm}
Layer Normalization (LN) \citep{ba2016layer} is widely adopted in Natural Language Processing, specially Transformer \citep{DBLP:journals/corr/VaswaniSPUJGKP17} incorporates LN as a standard normalization scheme. BN computes a mean and a variance over a mini-batch and the stability of training is highly dependent on these two statics. \citet{shen2020powernorm} has showed that transformer with BN leads to poor performance because of the large fluctuations of batch statistics throughout training. Layer normalization computes the mean and variance along the feature dimension for each training case. Different from BN, for each sample $h_i$, LN computes mean $\mu^{(L)}_{i}$ and variance $\sigma^{(L)}_{i}$ across the feature dimension. The normalization equations of LN are as follows:
\begin{equation}
\begin{split}
\hat{h}_{i} & = \gamma \odot \frac{h_i-\mu^{(L)}_i}{\sigma^{(L)}_i}+\beta, \\
\mu^{(L)}_i & =  \frac{1}{d}\sum_{j=1}^{d}h_{i_j}, \quad \sigma^{(L)}_{i} =\sqrt{\frac{1}{d}\sum_{j=1}^{d}(h_{i_j}-\mu^{(L)}_{i})^2},  
\end{split}
\end{equation}  
where $\hat{h_i}=[\hat{h}_{i_{1}},\hat{h}_{i_{2}},...,\hat{h}_{i_{d}}]$ is the feature-normalized response. $\gamma$ and $\beta$ are parameters with dimension $d$. 

Overall, there are many normalization approaches \citep{Ulyanov2016InstanceNT, Wu2018GroupN, shen2020powernorm, dimitriou2020new}. \citet{shen2020powernorm} has indicated that BN is suitable for computer vision tasks, while LN achieves better results on NLP. For a normalization approach, its performance may vary a lot in different tasks. So, it is very important to investigate the performance of normalization approaches in GNNs.   

\section{Datasets and experimental details}
\subsection{Dataset Statistics}
\label{appendix:statistics}

Table \ref{table:datasets} summarizes the statistics of the datasets used for our experiments.

\begin{table}[!h]
	\label{table:datasets}
	\centering
	\begin{tabular}{p{1.15cm}<{\centering}p{0.85cm}<{\centering}p{1cm}<{\centering}p{1.3cm}<{\centering}p{1.7cm}<{\centering}p{1.3cm}<{\centering}p{1cm}<{\centering}p{1cm}<{\centering}p{1cm}<{\centering}}
		\cmidrule[1.5pt](){1-9}
		\textbf{Dataset}  & \textbf{Graphs}   & \textbf{Nodes}  & \textbf{Total Nodes} & \textbf{Edges} & \textbf{Total Edges} &\textbf{Avg Edges} & \textbf{Task}  & \textbf{Classes}   \\
		\cmidrule[1.5pt](){1-9}
		PATTERN & 14K & 44-188 &166,449 & 752-14,864 &85,099,952 &51.1 & N.C. & 2 \\
		CLUSTER &  12K & 41-190 &140,643  &488-10,820 &51,620,680 &36.7 & N.C. & 6 \\
		SROIE     & 971  &18-153  &52,183       &70-2,031 & 420,903 & 8.1 & N.C. & 5 \\
		\cmidrule[0.5pt](){1-9}
		TSP         & 12K & 50-499 &3,309,140 &1,250-12,475 &82,728,500 & 25 & E.C. & 2 \\
		COLLAB  & 1     &235,868 & 235,868  &2,358,104 & 2,358,104 & 10 & E.C. & 2 \\
		\cmidrule[0.5pt](){1-9}
		MNIST     & 70K & 40-75 & 4,939,668 &320-600 & 39,517,344  & 8 & G.C. & 10 \\
		CIFAR10   & 60K & 85-150 &7,058,005 &680-1,200 &56,464,040 & 8 & G.C. & 10 \\
		ZINC      &12K   & 9-37     &277,864    &16-84   &597,970 & 2.1 & G.R. & - \\
		\cmidrule[1.5pt](){1-9}
	\end{tabular}
	\caption{Summary statistics of datasets used in our experiments. The $7$th column (AvgEdges) represents the average number of edges per node in a graph. N.C., E.C., G.C., G.R. mean node classification, edge classification, graph classification and graph regression independently.}
\end{table}

\subsection{SROIE}
\label{appendix:sroie}
For a receipt, each text bbox can be viewed as a node of a graph. Its positions, the attributes of bounding box, and the corresponding text are used as the node feature. To describe the relationships among all the text bounding boxes of a receipt, we consider the distance between two nodes $v_i$ and $v_j$. If the distance between two nodes is less than a threshold $\theta$, $v_i$ and $v_j$ are connected  by an edge $e_{i,j}$. The relative positions of two text bounding boxes are important for node classification, hence, we encode the relative coordinates of $v_i$ and $v_j$ to represent the edge $e_{ij}$. This information extraction problem can be treated as a node classification problem based on the graph. Our goal is to label each node (text bounding box) with five different classes, including Company, Date, Address, Total and Other. Since GatedGCN explicitly exploit edge features and has achieved state-of-the-art performance on various tasks, for this task, GatedGCN with 8 GCN layers is used. 

\begin{figure}[h]
	\centering
	\subfigure[A receipt image.]{
		\includegraphics[scale=0.16]{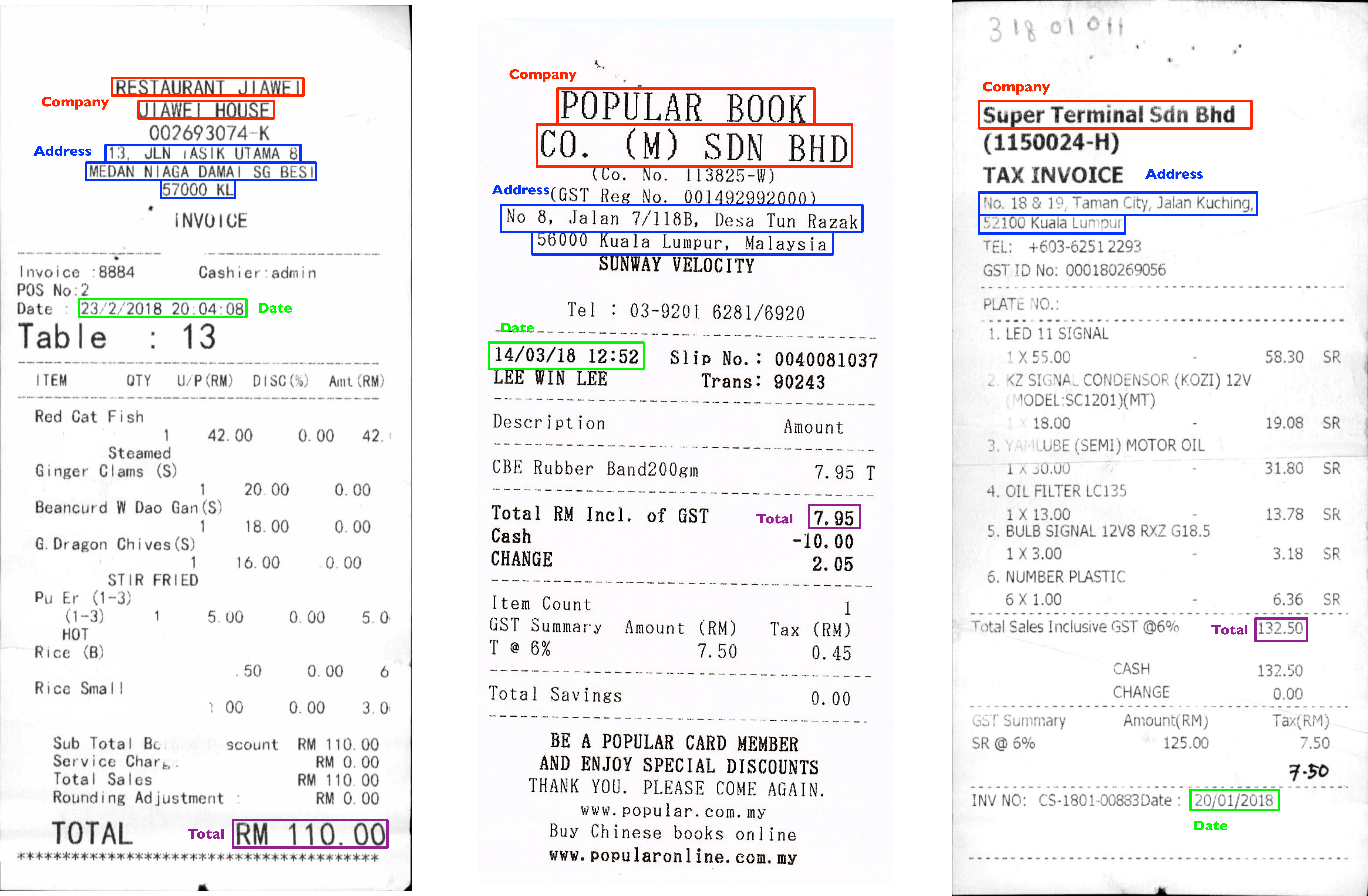}
	}
	\caption{Sample images of the SROIE dataset. Four entities are highlighted in different colors. ``Company'', ``Address'', ``Date'', and ``Total'' are marked with {\color{red}{Red}}, {\color{blue}{Blue}}, {\color{yellow}{Yellow}}, and {\color{purple}{Purple}} individually. The ``Company'' and the ``Address'' entities usually consist of several text lines.}
	\label{results_sroie}
\end{figure}

\section{Acknowledgement}
\label{appendix:acknowledgement}
We would like to thank Vijay~\etal \ to release their benchmarking code for our research. We also want to thank the DGL team for their excellent toolbox.

\end{document}